\pdfoutput=1

\documentclass[11pt]{article}

\usepackage{acl}

\usepackage{times}
\usepackage{latexsym}
\usepackage{mathtools}

\usepackage[T1]{fontenc}

\usepackage[utf8]{inputenc}

\usepackage{microtype}

\title{A Generative Model for Relation Extraction and Classification}

\author{Jian Ni, Gaetano Rossiello, Alfio Gliozzo, Radu Florian\\
    IBM Research AI\\
    1101 Kitchawan Road, Yorktown Heights, NY 10598, USA}

\begin{document}
\maketitle
\begin{abstract}
Relation extraction (RE) is an important information extraction task which provides essential information to many NLP applications such as knowledge base population and question answering. In this paper, we present a novel generative model for relation extraction and classification (which we call GREC), where RE is modeled as a sequence-to-sequence generation task. We explore various encoding representations for the source and target sequences, and design effective schemes that enable GREC to achieve state-of-the-art performance on three benchmark RE datasets. In addition, we introduce negative sampling and decoding scaling techniques which provide a flexible tool to tune the precision and recall performance of the model. Our approach can be extended to extract all relation triples from a sentence in one pass. Although the one-pass approach incurs certain performance loss, it is much more computationally efficient.
\end{abstract}

\section{Introduction}

Relation extraction (RE) is a fundamental information extraction task that seeks to detect and characterize semantic relationships between pairs of entities or events from natural language text. It provides important information for many NLP applications such as knowledge base population \cite{ji-grishman-2011-knowledge} and question answering \cite{xu-etal-2016-question}.

Relation extraction has been studied in two settings. In the first setting, gold entities are provided, and the RE task is to classify the relationships between given pairs of entities in sentences. This task is also known as \emph{relation classification} \cite{hendrickx-etal-2010-semeval,zhang-etal-2017-position}.

In the second setting, no gold entities are provided, and one needs to consider both entity recognition and relation extraction \cite{doddington-etal-2004-automatic}. This can be tackled via a \emph{pipeline} approach: first an entity recognition model is applied to extract entities \cite{florian-etal-2003-named,lample-etal-2016-neural}, and then a relation extraction model is applied to classify the relationships between all pairs of predicted entities \cite{kambhatla-2004-combining,chan-roth-2011-exploiting,zhong-chen-2021-frustratingly}. Alternatively, this can be addressed by a \emph{joint} approach where entity recognition and relation extraction are modeled and solved jointly \cite{li-ji-2014-incremental,miwa-bansal-2016-end,luan-etal-2019-general,lin-etal-2020-joint,wang-lu-2020-two}.

In this paper, we focus on RE with entities provided, which covers both the relation classification setting where gold entities are provided and the pipeline relation extraction setting where predicted entities are extracted via an entity recognition model. Given a sentence with a pair of gold or predicted entities, RE is naturally formulated as a classification task. It is a very challenging task since RE relies heavily on both syntactic and semantic information, with possibly multiple entities and relations existing in one sentence. 

\begin{figure*}
\centering
\includegraphics[scale=0.45]{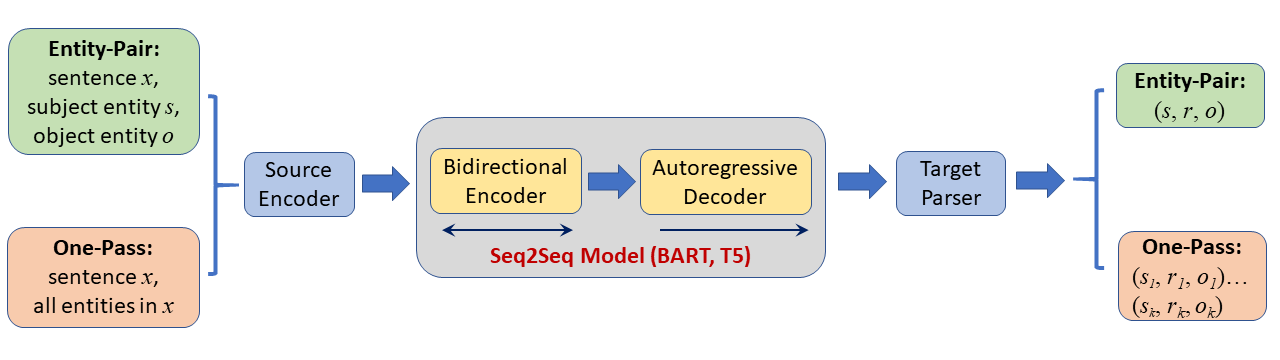} 
\caption{Overview of the Generative Relation Extraction and Classification (GREC) model.} \label{fig:genre-model}
\end{figure*}

We present a novel generative model for relation extraction and classification (named GREC), which treats RE as a sequence-to-sequence (seq2seq) text generation task. Given an input sentence and a pair of entities in it, the model generates an output \emph{relation triple} which consists of the two entities and a relation type that specifies their relationship. Compared with classification based RE approaches, the generative approach has the capability of encoding entity information in the target sequence. Experiment results show that GREC achieves state-of-the art performance on three benchmark RE datasets. Moreover, our approach enables the adoption of standard fine-tuning procedures with pre-trained seq2seq language models~\cite{wolf-etal-2020-transformers}, without the need of designing ad-hoc architectures, hence facilitating the deployment in information extraction systems.

While the idea of using seq2seq models for RE was studied before \cite{zeng-etal-2018-extracting,Zeng_Zhang_Liu_2020,Nayak_Ng_2020,zhang-etal-2020-minimize}, the previous works focused on end-to-end relation extraction that jointly extracts entities and relations from sentences. In this paper we focus on relation extraction and classification with entities provided, and we show that it is beneficial to encode entity information both in the source and in the target sequences to achieve the best performance. 

We summarize our main contributions as follows:
\begin{itemize}
    \item We explore various encoding representations for the source and target sequences, and design effective schemes that enable the GREC model to achieve state-of-the-art performance on three popular benchmark RE datasets: ACE05~\cite{Walker2006}, SemEval 2010 Task 8~\cite{hendrickx-etal-2010-semeval} and TACRED~\cite{zhang-etal-2017-position}. 
    
    \item We introduce negative sampling during training to improve the recall performance of the model. We also develop a novel decoding scaling scheme during inference to improve the precision performance. These together provide a flexible tool to tune both the precision and the recall performance of the model.
    
    \item We extend the approach to extract all relation triples from a sentence in one pass. Although the one-pass approach incurs certain performance loss, it significantly reduces the training and decoding time, as we show in Section~\ref{ExperimentsEntityPairVsOnePass}.
    
\end{itemize}

\section{Method}

For given pairs or all pairs of entities in a sentence, the RE task is to detect and characterize the relationships between those pairs of entities \cite{doddington-etal-2004-automatic,hendrickx-etal-2010-semeval,zhang-etal-2017-position}. 
We present a generative approach for RE, which treats RE as a seq2seq text generation task. Given an input sentence\footnote{Our approach can be extended to cross-sentential relation extraction if we let the input sequence covers multiple sentences.} with gold or predicted entities, we propose two approaches of encoding the entity information in the source sequence. With the \emph{entity-pair} approach, each time we encode one pair of entities in the source sequence; while with the \emph{one-pass} approach, we encoded all the entities in the source sequence. 

In Figure~\ref{fig:genre-model} we show the overview of our \textbf{G}enerative \textbf{R}elation \textbf{E}xtraction and \textbf{C}lassification (GREC) model. First, an input sentence and either i) a pair of entities (under the entity-pair approach) or ii) all the entities (under the one-pass approach) in the sentence are encoded to a source sequence ($src\_seq$) via a source encoding module. Then, the source sequence is passed to a seq2seq model which generates a target sequence ($tgt\_seq$) that includes the textual representation of the relation triple(s) supported by the input sentence. Finally, the generated target sequence is processed by a target parsing module to extract either i) a relation triple that consists of the subject entity, the object entity, and the relation type between the two entities (under the entity-pair approach) or ii) all the relation triples in the sentence (under the one-pass approach).

\subsection{Entity-Pair Approach}
Under the entity-pair approach, each time we encode the entity information of one pair of entities of an input sentence in the source sequence.

Let $\mathbf{x}=(x_1, x_2, ..., x_n)$ be an input sentence with $n$ tokens. Let $\mathbf{s} = (x_{s_1},...,x_{s_2})$ and $\mathbf{o} = (x_{o_1},...,x_{o_2})$ be a pair of entities in the input sentence $\mathbf{x}$, where $\mathbf{s}$ is the subject (head) entity and $\mathbf{o}$ is the object (tail) entity, with entity types $T_s$ and $T_o$, respectively.

Let $R=\{r_1, r_2, ..., r_K\}$ be the set of pre-defined relation types. We use a null relation type $r_0$ (e.g., $r_0$=``None'') to indicate that the two entities under consideration do not have a relationship belonging to one of the $K$ relation types.

 We have explored various schemes to encode the entity information in the source sequence. 
The first scheme is to use some special tokens to mark the start and end of the entities to encode entity location information as in  \cite{10.1145/3357384.3358119,baldini-soares-etal-2019-matching}:
\begin{eqnarray} \label{eq:src_encoder1}
    f_1(\mathbf{x}, \mathbf{s}, \mathbf{o})  & = & \big(x_1, ..., \$, x_{s_1},...,x_{s_2}, \$, ...,  \nonumber \\
         & & \&, x_{o_1},...,x_{o_2}, \& , ..., x_n\big) 
\end{eqnarray}

The second scheme is to use the entity type of an entity to mark the start and end of that entity, in order to encode both the entity location and entity type information in the source sequence as in \cite{DBLP:journals/corr/abs-2010-08652,zhong-chen-2021-frustratingly}:
\begin{eqnarray} \label{eq:src_encoder2}
    f_1(\mathbf{x}, \mathbf{s}, \mathbf{o})  & = & \big(x_1, ..., T_s, x_{s_1},...,x_{s_2}, T_s, ...,  \nonumber \\ 
         & & T_o, x_{o_1},...,x_{o_2}, T_o, ..., x_n\big)
\end{eqnarray}

In Equations (\ref{eq:src_encoder1}) and (\ref{eq:src_encoder2}) we assume that $\mathbf{s}$ appears before $\mathbf{o}$ (i.e., $1\leq s_1 \leq s_2 < o_1 \leq o_2 \leq n$). If $\mathbf{s}$ appears after $\mathbf{o}$, the positions of the two entities will be switched.

Next, to encode the \emph{direction} of a relation (i.e., to encode which of the two entities is the subject entity and which is the object entity), we create the following sub-sequence:
\begin{eqnarray}
    f_2(\mathbf{s}, \mathbf{o}) & = & [\mathbf{s} 
\; \# \; T_s \; , \; \mathbf{o} \; \# \; T_o]
\end{eqnarray}
so that the subject entity always appears before the object entity in this sub-sequence.

In addition to sub-sequences $f_1$ and $f_2$, we find that adding the list of relation types is helpful to the GREC model:
\begin{equation} \label{eq:relations}
    f_3(R) = [r_1 - r_2 - \cdot \cdot \cdot - r_K]
\end{equation}

The final encoding of the source sequence (i.e., the input to the seq2seq model) is the concatenation of the three sub-sequences:
\begin{eqnarray} \label{eq:entitypair-src}
src\_seq  =  f_1(\mathbf{x}, \mathbf{s}, \mathbf{o}) \oplus f_2(\mathbf{s}, \mathbf{o}) \oplus f_3(R) 
\end{eqnarray}

For the target sequence, we have also explored various choices. First we find that adding the subject and object entities to the target sequence (i.e., generating a relation triple) is better than generating the relation type only. Among the different orders of the relation triple that we have tried, we find that generating the relation triple with the order ``subject, relation, object'' is the most effective:
\begin{eqnarray} \label{eq:target_sro}
tgt\_seq & = &  [\mathbf{s}\; | \; r(\mathbf{s}, \mathbf{o}) \; | \; \mathbf{o}]
\end{eqnarray}
where $r(\mathbf{s}, \mathbf{o})$ is the relation type that specifies the relationship between the subject entity $\mathbf{s}$ and the object  entity $\mathbf{o}$. 

The seq2seq model is trained to generate a target sequence that contains the special characters `[', `]', and `|', which are used to parse the target sequence to a relation triple.  If $r(\mathbf{s}, \mathbf{o})=r_0$ (null relation type), the triple is a \emph{negative} example; otherwise, the triple is a \emph{positive} example.

Examples of the source and target sequence encoding are shown in Table~\ref{table:encoding_examples} (Appendix~\ref{sec:appendix}).

\subsubsection{Improving Recall via Sampling Negative Training Examples}

It can be very challenging to achieve a sufficient \emph{recall} for RE models \cite{zhang-etal-2017-position}. Let $R_{gold}$ be the set of gold positive relation triples in an RE dataset. Let $R_{pred}$ be the set of predicted positive relation triples of an RE model when applied on the dataset. The \emph{precision} and \emph{recall} of the RE model on the dataset are defined as:
\begin{eqnarray*}
    precision & = & \frac{|R_{pred} \cap R_{gold}|}{|R_{pred}|} \nonumber \\
    recall & = & \frac{|R_{pred} \cap R_{gold}|}{|R_{gold}|} \nonumber 
\end{eqnarray*}
where $|A|$ is the size (cardinality) of set $A$.

To improve the recall, one can try to let the RE model predict more positive relation triples to increase the number of true positives (a \emph{true positive} is a predicted positive relation triple that matches a gold positive relation triple, i.e., a triple in $|R_{pred} \cap R_{gold}|$). 

We find that sampling negative training examples during training is very effective for improving the recall of the GREC model. Specifically, we keep all the positive training examples while randomly sampling a fraction $\alpha$ of the total negative training examples for training the GREC model. $\alpha$ is called the \emph{negative sampling ratio} which is a number between 0 and 1. As we decrease $\alpha$, the model is trained with fewer negative examples and higher positive-to-negative ratio, and it will generate more positive relation triples during inference, hence improving the recall. We observe that sampling negative training examples, however, might reduce the precision. In the next subsection we present a scheme to improve the precision.

\subsubsection{Improving Precision via Decoding Scaling}

When the GREC model generates more positive relation triples and gets more true positives, the recall can be improved. However, this may also increase the number of false positives and reduce the precision. We propose a novel \emph{decoding scaling} scheme that utilizes the sequence scores of the top $N$ generated target sequences to improve the precision.

For an input source sequence $\mathbf{z}$, we let the GREC model generate top $N$ target sequences (relation triples) $\mathbf{y}_1, ..., \mathbf{y}_N$ with the highest sequence scores, where the \emph{sequence score} of $\mathbf{y}_i$ is the conditional probability of $\mathbf{y}_i$ given  $\mathbf{z}$: $P(\mathbf{y}_i | \mathbf{z})$. Note that in normal decoding, we just let the GREC model generate the best target sequence $\mathbf{y}_1$ and use that as the prediction.

If a relation triple includes a non-null relation type in $R$, we call it a positive triple; otherwise we call it a negative triple. There are two cases to consider:
\begin{itemize}
\item[(1)] If the top $N$ triples are all positive or all negative, the scheme simply selects the best positive or negative triple with the highest sequence score. This is the same as in normal decoding. 

\item[(2)] If the top $N$ triples include both positive and negative triples, let $\mathbf{y}_+^*$ and $\mathbf{y}_-^*$ be the best positive and negative triple, respectively. We select the triple $\mathbf{y}^*$ as the prediction as follows:
\begin{equation}
    \mathbf{y}^*= 
\begin{dcases}
    \mathbf{y}_+^*,& \text{if } \frac{P(\mathbf{y}_+^* | \mathbf{z})}{P(\mathbf{y}_-^* | \mathbf{z})}\geq \beta\\
    \mathbf{y}_-^*,              & \text{otherwise}
\end{dcases}
\end{equation}

\end{itemize}
where $\beta$ is called the \emph{decoding scaling factor}. When $\beta=1$ (no scaling), the scheme will just select the best generated triple as in normal decoding. When $\beta > 1$, the scheme will select the best positive triple $\mathbf{y}_+^*$ only if its score is greater than the score of the best negative triple $\mathbf{y}_-^*$ by a margin, so the predicted positive triple is more likely to be a true positive.  Therefore, the total number of false positives will be reduced, hence improving the precision.

\subsection{One-Pass Approach}

Under the one-pass approach, each time we encode the information of all the entities of the input sentence $\mathbf{x}$ in the source sequence. The target sequence also includes all the positive relation triples supported by the input sentence.

Let $E(\mathbf{x})$ be the set of all entities in $\mathbf{x}$, where an entity $\mathbf{e}_i = (x_{i_1},...,x_{i_2}) \in E(\mathbf{x})$ is a span in $\mathbf{x}$, with entity type $T_i$. Let $
R(\mathbf{x})$ be the set of all positive relation triples supported by $\mathbf{x}$, where a relation triple $\mathbf{t}_j = (\mathbf{s}_j, r_j, \mathbf{o}_j) \in R(\mathbf{x})$ consists of a subject entity $\mathbf{s}_j \in E(\mathbf{x}) $, an object entity $\mathbf{o}_j \in E(\mathbf{x})$, and their relation type $r_j \in R$.

First we extend the entity type marking to all the entities in the input sentence as follows:
\begin{eqnarray} \label{eq:entity-type-marker-input}
   &     & f_1(\mathbf{x}, E(\mathbf{x})) \nonumber \\
   &  =  & \Big(x_1, ..., T_i, x_{i_1},...,x_{i_2}, T_i, ..., x_n, \forall \mathbf{e}_i \in E(\mathbf{x}) \Big) \nonumber \\
\end{eqnarray}

Then we encode the list of entities with their entity types as follows:
\begin{equation}
       f_2(E(\mathbf{x})) = [\mathbf{e}_i 
\; \# \; T_i \; , \forall \mathbf{e}_i \in E(\mathbf{x})]
\end{equation}

We also include the list of relation types as in (\ref{eq:relations}). The final encoding of the source sequence is:
\begin{equation}
    src\_seq = f_1(\mathbf{x},E(\mathbf{x}) ) \oplus f_2(E(\mathbf{x})) \oplus f_3(R)
\end{equation}

The encoding of the target sequence is:
\begin{equation} \label{eq:one-pass-target}
    tgt\_seq = \oplus_{(\mathbf{s}_j, r_j, \mathbf{o}_j) \in R(\mathbf{x})}[\mathbf{s}_j\; | \; r_j \; | \; \mathbf{o}_j]
\end{equation}
In case there is no positive relation triple in $\mathbf{x}$ (i.e., $R(\mathbf{x})=\emptyset$), we set $tgt\_seq$ to be ``[None | None | None]''.

If no entity information is provided, then the encoding of the source sequence is the concatenation of the input sentence and the list of relation types:
\begin{equation}
    src\_seq = \mathbf{x} \oplus f_3(R)
\end{equation}
The encoding of the target sequence is the same as in (\ref{eq:one-pass-target}).

Note that for an input sentence with $m$ entities, the entity-pair approach will create $m(m-1)$ $src\_seq$ and $tgt\_seq$ pairs, while the one-pass approach will create just one $src\_seq$ and $tgt\_seq$ pair, which is much more computationally efficient.

\subsection{Seq2Seq Model}

For both the entity-pair approach and the one-pass approach, we apply a seq2seq model such as BART \cite{lewis-etal-2020-bart} or T5 \cite{2020t5} to convert a source sequence to a target sequence. BART/T5 uses a standard Transformer based neural machine translation architecture \cite{Vaswani2017} with multiple bidirectional encoder layers and autoregressive decoder layers. Both models were pre-trained with large English text corpora as a denoising autoencoder that maps a corrupted document to the original document. We convert the RE data to source and target sequence pairs as described in the previous subsections, and use them to fine-tune the seq2seq models for relation triple generation.

\section{Experiments}

\subsection{Datasets}
We evaluate the GREC model on 3 popular benchmark relation extraction and classification datasets: ACE05, SemEval 2010 Task 8, and TACRED.

The ACE05 dataset \cite{Walker2006} is a benchmark relation extraction dataset developed by the Linguistic Data Consortium for the purpose of Automatic Content Extraction (ACE) technology evaluation. ACE05 defines 7 entity types and 6 relation types between the entities. 
We use the same training, development, and test data split in prior works \cite{li-ji-2014-incremental,luan-etal-2019-general}.

The SemEval 2010 Task 8 dataset \cite{hendrickx-etal-2010-semeval} is a benchmark dataset for relation classification.  It defines 9 relation types and a null relation type ``Other''. It has 8000 training examples and 2717 test examples. We randomly select 1000 training examples for development.

TACRED~\cite{zhang-etal-2017-position} is a large supervised relation classification dataset obtained via crowdsourcing. It defines 42 relation types (including a null relation type ``no\_relation'') and includes over 100K examples. The dataset was recently revised and improved in \cite{alt-etal-2020-tacred} by reducing the annotation errors. In our experiments we use this revised version, which includes 68,124 training examples, 22,631 development examples, and 15,509 test examples.

\subsection{Implementation Details}

We build the GREC model on top of Transformer based seq2seq models including BART and T5, with HuggingFace's pytorch implementation \cite{wolf-etal-2020-transformers}. Our preliminary experiment results (Table~\ref{tab:ablation_pretrained}) show that BART and T5 achieve similar performance. We choose one model (bart-large) to run all the experiments, which is a common practice to reduce the total computational cost and energy consumption. 

We use the development sets to tune the hyper-parameters. We learn the model parameters using Adam \cite{KingmaBa2015}, with a learning rate $l=3e$-5, a training batch size of $b=16$ for ACE05 and SemEval 2010 Task 8, and $b=8$ for TACRED. We train the GREC model for 10 epochs with the entity-pair approach and 20 epochs with the one-pass approach. All experiments were conducted on a 2 NVIDIA V100 GPUs computer.

\begin{table*}
    \centering
    \begin{tabular}{r|c|c|c}
         \textbf{Model} & \textbf{Entity} & \textbf{Rel} & \textbf{Rel+}   \\\hline
         \cite{li-ji-2014-incremental} & 80.8 & 52.1 & 49.5 \\
         SPTree \cite{miwa-bansal-2016-end} & 83.4 & - & 55.6 \\
         \cite{katiyar-cardie-2017-going} & 82.6 & 55.9 & 53.6\\
         \cite{zhang-etal-2017-end} & 83.6 & - & 57.5 \\
         MRT \cite{sun-etal-2018-extracting} & 83.6 & - & 59.6 \\
         \cite{li-etal-2019-entity} & 84.8 & - & 60.2 \\
         \cite{dixit-al-onaizan-2019-span} & 86.0 & - & 62.8 \\
         DYGIE \cite{luan-etal-2019-general}$^*$ & 88.4 & 63.2 & - \\
         DyGIE++ \cite{wadden-etal-2019-entity}$^*$ & 88.6 & 63.4 & - \\
         \cite{lin-etal-2020-joint} & 88.8 & 67.5 & - \\
         \cite{wang-lu-2020-two} & 89.5 & 67.6 & 64.3 \\
          
          TANL \cite{tanl} & 88.9 & 63.7 & - \\
         PURE - single sentence \cite{zhong-chen-2021-frustratingly} & 89.7 & 69.0 & 65.6 \\
          PURE - cross sentence \cite{zhong-chen-2021-frustratingly}$^*$ & 90.9 & 69.4 & 67.0 \\
         GREC (ours) & 90.4 & \textbf{70.2} $\pm$ 0.4 &  \textbf{68.2} $\pm$ 0.5\\ \hline
         
    \end{tabular}
    \caption{Micro $F_1$ scores on the ACE05 test set. For GREC we report the \emph{mean} and \emph{standard deviation} of the performance over 5 runs. $^*$These models use cross-sentence information. }
    \label{tab:ace05_entitypair}
\end{table*}

\begin{table*}
    \centering
    \begin{tabular}{r|c}
         \textbf{Model} &  \textbf{Macro} $\mathbf{F_1}$  \\\hline
         CNN \cite{zeng-etal-2014-relation} & 82.7 \\
         Attention Bi-LSTM  \cite{zhou-etal-2016-attention} & 84.0 \\
         CR-CNN  \cite{dos-santos-etal-2015-classifying} & 84.1 \\
         Bi-LSTM \cite{zhang-etal-2015-bidirectional} & 84.3 \\
         Hierarchical Attention RNN \cite{xiao-liu-2016-semantic} & 84.3 \\
         Entity Attention Bi-LSTM \cite{DBLP:journals/corr/abs-1901-08163} & 85.2 \\
         Attention CNN \cite{shen-huang-2016-attention} & 85.9 \\
         TRE \cite{DBLP:conf/akbc/AltHH19} & 87.1 \\
         SpanRel \cite{jiang-etal-2020-generalizing} & 87.4 \\
         Multi-Attention CNN  \cite{wang-etal-2016-relation} & 88.0 \\
         KnowBERT-W+W~\cite{peters-etal-2019-knowledge}$^*$ & 89.1 \\
         R-BERT \cite{10.1145/3357384.3358119} & 89.25 \\
         BERT$_{EM}$ \cite{baldini-soares-etal-2019-matching} & 89.2 \\
         BERT$_{EM}$+MTB~\cite{baldini-soares-etal-2019-matching}$^*$ & 89.5 \\
         GREC (ours) & \textbf{89.9} $\pm$ 0.1 \\
\hline
    \end{tabular}
    \caption{Macro $F_1$ scores on the SemEval 2010 Task 8 test set. For GREC we report the \emph{mean} and \emph{standard deviation} of the performance over 5 runs.  $^*$These models use additional data derived from Wikipedia/WordNet to pre-train their models.}
    \label{tab:semeval_entitypair}
\end{table*}

\begin{table*}
    \centering
    \begin{tabular}{r|c}
         \textbf{Model} & \textbf{Micro} $\mathbf{F_1}$ \\\hline
         LSTM (masked) ~\cite{zhang-etal-2017-position} & 63.9 \\
         LSTM + BERT (masked)~\cite{alt-etal-2020-tacred} & 73.4  \\
         CNN (masked)~\cite{nguyen-grishman-2015-relation} & 66.5 \\
         CNN + BERT (masked)~\cite{alt-etal-2020-tacred} & 74.3 \\
         TRE~\cite{DBLP:conf/akbc/AltHH19} & 75.3 \\
         SpanBERT~\cite{joshi-etal-2020-spanbert} & 78.0 \\
         KnowBERT-W+W~\cite{peters-etal-2019-knowledge}$^*$ & 79.3 \\
         GREC (ours) & \textbf{80.6} $\pm$ 0.6 \\ \hline
    \end{tabular}
    \caption{Micro $F_1$ scores on the TACRED-Revised test set. For GREC we report the \emph{mean} and \emph{standard deviation} of the performance over 5 runs. $^*$This model was pre-trained with additional data derived from Wikipedia and WordNet.}
    \label{tab:tacred_entitypair}
\end{table*}

\subsection{Main Results}

Our best GREC model with the entity-pair approach uses source sequence encoding (\ref{eq:entitypair-src}) with entity type markers (\ref{eq:src_encoder2}) and target sequence encoding (\ref{eq:target_sro}). The results of the GREC model reported in Tables~\ref{tab:ace05_entitypair}-\ref{tab:tacred_entitypair} include the \emph{mean} and \emph{standard deviation} of the performance over 5 runs with different random seeds.

In Table~\ref{tab:ace05_entitypair} we compare the GREC model with previous approaches on the ACE05 test set. As in prior works we use micro-averaged $F_1$ score as the evaluation metric. For entity recognition, a predicted entity is considered correct if its predicted entity span and entity type are both correct.  For relation extraction with predicted entities, following \cite{li-ji-2014-incremental,wang-lu-2020-two,zhong-chen-2021-frustratingly},  we use two evaluation metrics: 1) \emph{Rel}: a predicted relation is considered correct if the two predicted entity spans and the predicted relation type are correct;  2) \emph{Rel+}: a predicted relation is considered correct if the two predicted entity spans and entity types as well as the predicted relation type are all correct. Our entity recognition model is an ensemble of RoBERTa \cite{DBLP:journals/corr/abs-1907-11692} based sequence labeling models with voting.

As shown in Table~\ref{tab:ace05_entitypair}, the GREC model achieves the state-of-the-art performance on ACE05. GREC improves the previous best model PURE by 1.2 $F_1$ points on the Rel metric and by 2.6 $F_1$ points on the Rel+ metric with the single-sentence setting.

\begin{table}
    \centering
    \begin{tabular}{ccccc}
    \hline
        \textbf{Source} & \textbf{Target} &  $P$ & $R$ & $F_1$ \\ \hline
          no entity marker & $[s | r | o]$ &  70.2 & 68.8 & 69.5 \\
          special token marker & $[s | r | o]$ & 73.2  & 68.2 & 70.6  \\
          \textbf{entity type marker} & $\mathbf{[s | r | o]}$  & \textbf{74.4} & \textbf{76.1} & \textbf{75.2} \\
          entity type marker & $[r]$  & 71.2 & 66.9 & 69.0 \\
          entity type marker & $[r | s | o]$  & 71.8 & 75.3 & 73.5 \\
          entity type marker & $[s | o | r]$  & 73.7 & 72.8 & 73.3 \\

    \end{tabular}
    \caption{Performance (precision, recall, $F_1$ score) of the GREC model on the ACE05 development set (with gold entities) under different source and target sequence encoding representations.}
    \label{tab:ablation_encoding}
\end{table}

\begin{table*}
    \centering
    \begin{tabular}{c|ccc|ccc|ccc|ccc}
    
           & \multicolumn{3}{c|}{$\beta$ = 1} & \multicolumn{3}{c|}{$\beta$ = 1.1} & \multicolumn{3}{c|}{$\beta$ = 1.2} & \multicolumn{3}{c}{$\beta$ = 1.3} \\
           & $P$ & $R$ & $F_1$  & $P$ & $R$ & $F_1$ & $P$ & $R$ & $F_1$ & $P$ & $R$ & $F_1$ \\ \hline
           $\alpha=1$  & 72.1 & 75.0 &  73.5 & 74.2 & 73.0 & 73.6 & 76.2 & 71.4 & 73.8 & \textbf{77.5} & 69.0 & 73.0 \\ \hline
           $\alpha=0.9$  & 71.6 & 78.7 & 75.0 & 73.7 & 75.9 & 74.7 & 76.6 & 73.3 & 74.9 & \textbf{77.5} & 70.4 & 73.8 \\ \hline
           $\alpha=0.8$ & 68.2 & \textbf{79.3} & 73.4 & 71.7 & 78.2 & 74.8 & 74.4 & 76.1 & \textbf{75.2} & 76.4 & 73.3 & 74.8 \\ \hline
           $\alpha=0.7$ & 66.6 & 78.7 & 72.2 & 69.9 & 76.9 & 73.2 & 72.1 & 74.6 & 73.3 & 74.5 & 72.5 & 73.5 \\ \hline
          
    \end{tabular}
    \caption{Performance of the GREC model on the ACE05 development set (with gold entities) under different negative sampling ratio $\alpha$ and decoding scaling factor $\beta$.}
    \label{tab:ablation_alpha_beta}
\end{table*}

\begin{table}
    \centering
    \begin{tabular}{ccc}
    \hline
        \textbf{Model} &  \textbf{Parameters}  & $F_1$ \\ \hline
        t5-base    & $L$=12, $A$=12, $H$=768  & 72.6 \\
        bart-base  & $L$=12, $A$=16, $H$=768  & 73.5 \\
        bart-large & $L$=24, $A$=16, $H$=1024 & 75.2 \\

    \end{tabular}
    \caption{Performance ($F_1$ score) of the GREC model on the ACE05 development set (with gold entities) under different pre-trained seq2seq models. $L$ is the number of transformer layers, $A$ is the number of attention heads, and $H$ is the hidden state vector size.}
    \label{tab:ablation_pretrained}
\end{table}

\begin{table*}[h]
    \centering
    \begin{tabular}{cc|ccc|cc}
           \multicolumn{2}{c}{\textbf{Setup}}  &  \multicolumn{3}{|c}{\textbf{Performance}} & \multicolumn{2}{|c}{\textbf{Computational Cost}}  \\\hline
           Approach &  Entities  & $P$ & $R$ & $F_1$ & Training & Decoding  \\\hline
          entity-pair & \emph{gold}  & 74.4 & 76.1 & 75.2 & 559 mins & 459 secs  \\ 
          entity-pair & \emph{predicted}  & 66.7 & 68.3 & 67.5 & 559 mins & 456 secs  \\ \hline
          one-pass &  \emph{gold}  & 66.0 & 65.8  & 65.9 & 90 mins & 32 secs  \\ 
          one-pass & \emph{predicted} & 63.4 & 56.5 & 59.7 & 90 mins & 28 secs \\ \hline
          one-pass & \emph{no}  & 55.8 & 51.9  & 53.8 & 80 mins & 29 secs \\ \hline
    \end{tabular}
    \caption{Performance and computational cost of the GREC model under the entity-pair approach and the one-pass approach on the ACE05 development set. Training and decoding time is measured on a 2 NVIDIA V100 GPUs computer with a batch size of 16, 10 training epochs for entity-pair and 20 training epochs for one-pass. \emph{gold}: the gold entities are given during inference. \emph{predicted}: the predicted entities are given during inference. \emph{no}: no entities are given during inference (and training).}
    \label{tab:onepass_results}
\end{table*}

In Table~\ref{tab:semeval_entitypair} we compare the GREC model with previous approaches on the SemEval 2010 Task 8 test set. As in prior works we use the SemEval 2010 Task 8 official scoring metric which is macro-averaged
$F_1$ score for the 9 relation types (excluding the null relation type ``Other'') and takes directionality into account. The GREC model achieves the state-of-art performance. While the BERT$_{EM}$+MTB model used additional data (600 million relation statement pairs derived from English Wikipeida) to pre-train the model, GREC achieves better performance without using any additional data. 

In Table~\ref{tab:tacred_entitypair} we compare the GREC model with previous approaches on the revised TACRED test set. As in prior works we use micro-averaged $F_1$ score as the evaluation metric. Again, the GREC model achieves the state-of-the-art performance without using any additional data.

\subsection{Ablation Studies}

In this subsection we study the contributions of different components on the GREC model.

\subsubsection{Source and Target Sequence Encoding}

In Table~\ref{tab:ablation_encoding} we show the performance of the GREC model on the ACE05 development set  under different source and target sequence encoding representations. There are two observations:
\begin{itemize}
    \item For the source sequence encoding, it is important to encode the entity information in the input sentence using entity markers. The special token markers (\ref{eq:src_encoder1}) that encode the entity location information improved the performance by 1.1 $F_1$ points, and the entity type markers (\ref{eq:src_encoder2}) that encode both the entity location and type information improved the performance by 5.7 $F_1$ points, compared with not using any entity markers.
    
    \item For the target sequence encoding, it is beneficial to add the subject and object entities in the target sequence, which helps the GREC model to generate more accurate relation types. This improved the performance by 4+ $F_1$ points compared with generating the relation type only ($[r]$). Among the different orders of the relation triple that we have tried, the order ``subject, relation, object'' ($[s|r|o]$) achieved the best performance. The reason could be that the order ``subject, relation, object'' (e.g., ``Toefting, works for, Bolton'') is the one that is the most consistent with the English language SVO order and the seq2seq model (BART) was pre-trained with English text.
    
\end{itemize}

\subsubsection{Negative Sampling and Decoding Scaling}

In Table~\ref{tab:ablation_alpha_beta} we show the performance of the GREC model on the ACE05 development set under different negative sampling ratio $\alpha$ and decoding scaling factor $\beta$ (we let the GREC model generate top $N$=5 target sequences). The key observations are:
\begin{itemize}
    \item For a fixed decoding scaling factor $\beta$ (a column in Table~\ref{tab:ablation_alpha_beta}), as we decrease the negative sampling ratio $\alpha$ (i.e., keep fewer negative training examples during training), the recall is improved. The recall reached the highest value at $\alpha=0.8$, and further decreasing $\alpha$ could reduce the recall.
    
    \item For a fixed negative sampling ratio $\alpha$ (a row in Table~\ref{tab:ablation_alpha_beta}), as we increase the decoding scaling factor $\beta$ (so the predicted positive triple is more likely to be a true positive triple), the precision is improved. However, increasing $\beta$ hurts the recall. 
    
    \item Negative sampling and decoding scaling provide a flexible tool to tune the precision and recall performance of the GREC model. If we want to achieve a high recall, we would keep $\beta=1$ (no decoding scaling) and select an optimal $\alpha$: in this case $\alpha=0.8$ gives the best recall performance of 79.3. On the other hand, if we want to have a higher precision, we would keep $\alpha=1$ (no negative sampling) and pick a larger $\beta$.  We can also use the development set to find the optimal $\alpha$ and $\beta$ that achieve the highest $F_1$ score.
\end{itemize}

\subsubsection{Pre-trained Seq2Seq Models}

In Table~\ref{tab:ablation_pretrained} we show the performance of the GREC model under 3 pre-trained seq2seq models: t5-base, bart-base and bart-large. The performance of GREC is robust across different pre-trained models, with the deeper and larger bart-large model achieves the best performance.

\subsection{Entity-Pair vs. One-Pass Approach} \label{ExperimentsEntityPairVsOnePass}

A sentence can have multiple entities and relation triples in the ACE05 data, so we use ACE05 to compare the performance and computational cost of the GREC model under the entity-pair approach and the one-pass approach.

As shown in Table~\ref{tab:onepass_results}, the entity-pair approach has a clear advantage over the one-pass approach on performance (nearly 10 $F_1$ points gain). On the other hand, since the one-pass approach creates just one source sequence for a sentence and extracts all the relation triples from the sentence in one-pass, it has a much smaller number of training/test examples and hence lower computational cost (6x faster for training and 15x faster for decoding) compared with the entity-pair approach. Another key observation is that encoding entity information (even predicted) can significantly improve the performance compared with no entities provided.

\section{Related Work}

Many RE models have been developed to improve the performance on benchmark RE datasets such as ACE05, SemEval 2010 Task 8 and TACRED. Earlier RE models require extensive feature engineering to derive and combine various lexical, syntactic and semantic features \cite{kambhatla-2004-combining,zhou-etal-2005-exploring,chan-roth-2011-exploiting,li-ji-2014-incremental}. Later neural network based RE models have become dominant, including CNN based models \cite{zeng-etal-2014-relation,dos-santos-etal-2015-classifying,nguyen-grishman-2015-relation}, RNN based models \cite{zhang-etal-2015-bidirectional,xiao-liu-2016-semantic,miwa-bansal-2016-end,ni-florian-2019-neural}, and most recently Transformer based models \cite{10.1145/3357384.3358119,baldini-soares-etal-2019-matching,zhong-chen-2021-frustratingly}.

Seq2seq models have been used for NLP tasks such as machine translation~\cite{10.5555/2969033.2969173,cho-etal-2014-learning} and text summarization~\cite{rush-etal-2015-neural,chopra-etal-2016-abstractive}. Recently, generative approaches based on seq2seq models have been proven competitive in NLP applications such as question answering, fact checking,  relation linking and intent classification~\cite{DBLP:conf/nips/LewisPPPKGKLYR020,petroni-etal-2021-kilt,DBLP:conf/semweb/RossielloMABGNK21,ahmad-etal-2021-intent}. 
While seq2seq models were also applied to RE \cite{zeng-etal-2018-extracting,Zeng_Zhang_Liu_2020,Nayak_Ng_2020,zhang-etal-2020-minimize,tanl}, the previous works focused on end-to-end relation extraction that jointly extracts entities and relations from sentences. Our work is also based on the seq2seq framework. The main difference is that we focus on relation  extraction and classification with entities provided, and we show it is beneficial to encode entity information both in the source and target sequences to achieve the best performance.

\section{Conclusion}

In this paper we presented a novel generative model for relation extraction and classification. We showed the importance of encoding entity information in the source and target sequences and designed effective encoding representations that enable the model to achieve state-of-the-art performance on three popular benchmark RE datasets. Our model is easy to implement with standard pre-trained seq2seq models like BART, has components to flexibly tune the precision and recall performance, and has the potential of significantly reducing the training and decoding time via the one-pass formulation.

\bibliography{anthology,custom}
\bibliographystyle{acl_natbib}

\appendix
\section{Examples of Source and Target Sequence Encoding} \label{sec:appendix}

\begin{table*}
\begin{center}
\begin{tabular}{|c|p{8.5cm}|p{4.5cm}|}

\hline
Approach & Source Sequence & Target Sequence  \\ 
\hline
Entity-Pair & \texttt{\textbf{Person} Toefting \textbf{Person} transferred to \textbf{Organization} Bolton \textbf{Organization} in February 2002 from German club Hamburg. [Toefting \# Person , Bolton \# Organization] [affiliated to - located at - makes - part of - relationship - works for]} & \texttt{[Toefting | works for | Bolton]}   \\

\hline

One-Pass & \texttt{\textbf{Person} Toefting
 \textbf{Person} transferred to \textbf{Organization} Bolton \textbf{Organization} in February 2002 from \textbf{Geo-political} German \textbf{Geo-political} \textbf{Organization} club \textbf{Organization} \textbf{Organization}
Hamburg \textbf{Organization}. [Toefting \# Person , Bolton \# Organization , German \# Geo-political , club \# Organization , Hamburg \# Organization] [affiliated to - located at - makes - part of - relationship - works for]} & \texttt{[Toefting | works for | Bolton] [Toefting | works for | Hamburg] [club | affiliated to |
German]}  \\
\hline

\end{tabular}
\end{center}

\caption{Examples of the source and target sequence encoding under the entity-pair approach and the one-pass approach. For the source sequence, entity type markers are shown in bold.} \label{table:encoding_examples}
\end{table*}


\end{document}